\documentclass[sigconf,natbib=true]{acmart}
\AtBeginDocument{%
  \providecommand\BibTeX{{%
    \normalfont B\kern-0.5em{\scshape i\kern-0.25em b}\kern-0.8em\TeX}}}

\begin{document}

\title{NC-DRE: Leveraging Non-entity Clue Information for Document-level Relation Extraction}

\author {
    Liang Zhang\textsuperscript{\rm 1} 
    Yidong Cheng \textsuperscript{\rm 1}\\
    \textsuperscript{\rm 1} Department of Artificial Intelligence, School of Informatics, Xiamen University\\
    \texttt{ \{lzhang,ydcheng\}@stu.xmu.edu.cn} \\
}

\renewcommand{\shortauthors}{Trovato and Tobin, et al.}

\begin{abstract}
Document-level relation extraction (RE), which requires reasoning on multiple entities in different sentences to identify complex inter-sentence relations, is more challenging than sentence-level RE. 
To extract the complex inter-sentence relations, previous studies usually employ graph neural networks (GNN) to perform inference upon heterogeneous document-graphs.
Despite their great successes, these graph-based methods, which normally only consider the words within the mentions in the process of building graphs and reasoning, tend to ignore the non-entity clue words that are not in the mentions but provide important clue information for relation reasoning.
To alleviate this problem, we treat graph-based document-level RE models as an encoder-decoder framework, which typically uses a pre-trained language model as the encoder and a GNN model as the decoder, and propose a novel graph-based model \textbf{NC-DRE} that introduces decoder-to-encoder attention mechanism to leverage \textbf{N}on-entity \textbf{C}lue information for \textbf{D}ocument-level \textbf{R}elation \textbf{E}xtraction. 
Moreover, to further improve the reasoning ability of NC-DRE, we utilize a variant of Transformer's decoder rather than the GNN models as the decoder.
Specifically, in the variant of Transformer's decoder, the standard Multi-head Self-Attention (MSA) mechanism is replaced with a new Structured-Mask Multi-head Self-Attention (\textbf{SM-MSA}) one.
The experiments on three public document-level RE datasets show that NC-DRE can effectively capture non-entity clue information during the inference process of document-level RE and achieves state-of-the-art performance.
\end{abstract}




\maketitle

\section{Introduction}

Human knowledge can be expressed and stored in the form of triples, and this structured knowledge is beneficial to many downstream tasks, such as question answering \cite{c:102}, knowledge graph completion \cite{c:101}, and search engine \cite{c:103}. Therefore, many efforts have been devoted to researching relation extraction (RE), which aims at extracting relational facts from plain text \cite{c:104}.
Although the early research on RE mostly focused on sentence-level RE, many recent efforts have extended RE to the document-level \cite{c:105,c:106}.
Moreover, since large amounts of relationships, such as relational facts from Wikipedia articles and biomedical literature, are expressed by multiple sentences  \cite{c:105,c:107}, document-level RE is more practical than sentence-level RE.

\begin{figure}[t]
\centering
\includegraphics[width=0.90 \columnwidth]{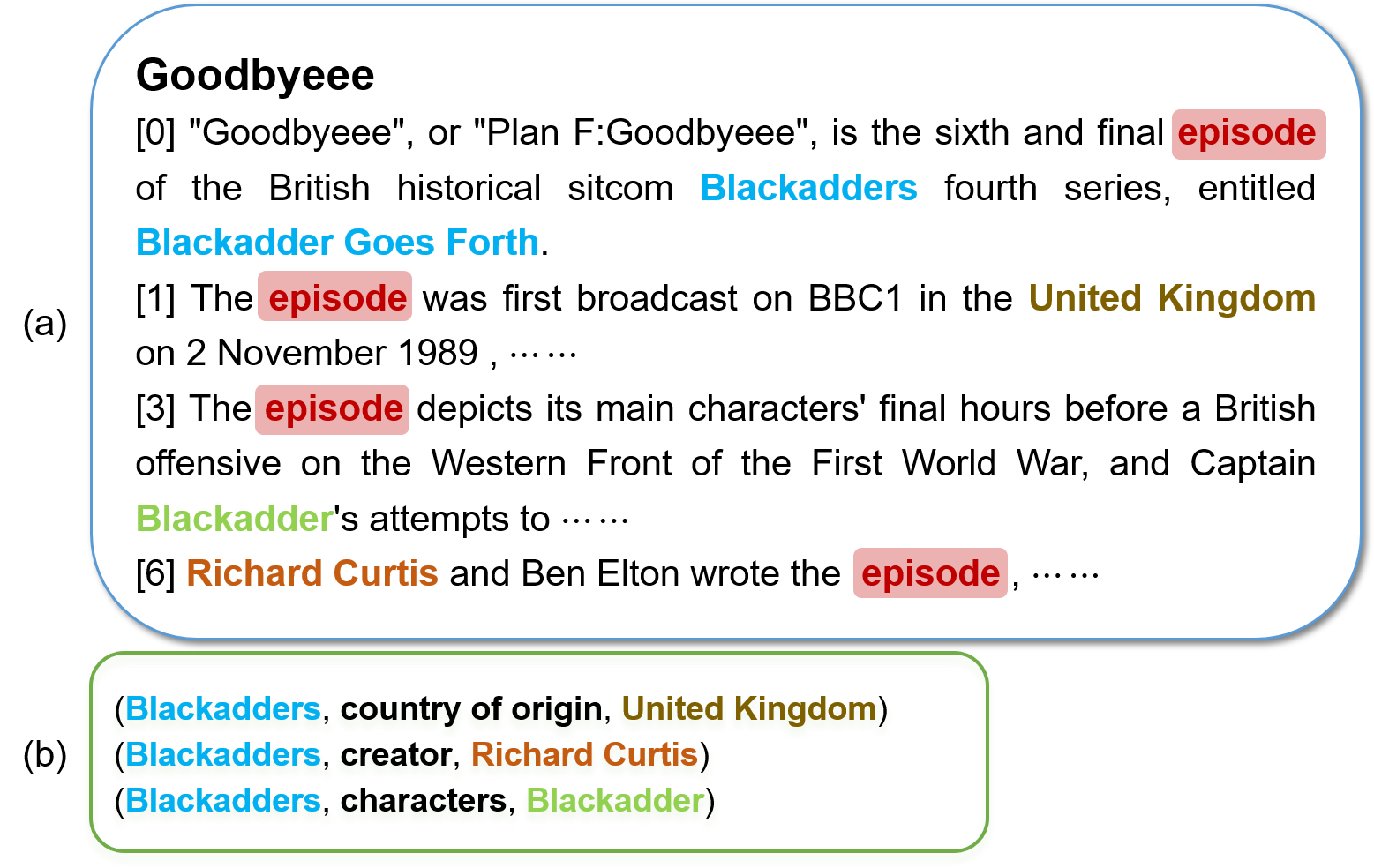} 
\caption{An example from the DocRED dataset that illustrates the importance of the clue words outside the mentions for relationship prediction. (a) is the document, in which different colors indicate different entities. And, the highlighted word \textit{episode} does not belong to any entity mentions but provides important clue information to predict the inter-sentence relations in (b).}
\label{fig1}
\end{figure}

Compared with sentence-level RE, document-level RE has one major challenge: the identification of many complex inter-sentence relations requires techniques of reasoning.
These inter-sentence relations may not be explicitly expressed in the document, which could only be extracted through reasoning techniques. 

Since graph neural networks (GNN) can not only effectively model long-distance dependencies but also has excellent inference ability, methods based on graphs are widely used for document-level RE \cite{c:129,c:130,c:133,c:131}.
These methods first obtain the contextual embedding of each word in the document through an encoder (e.g., a pre-trained language model), and then apply a pooling operation on on the words contained in the mentions (mention words) to gets the contextual embeddings of mentions or entities.
Finally, they construct heterogeneous document-graphs with dependency structures, heuristics, or structured attention \cite{c:108,c:109,c:110}, and perform inference on the document-graph through graph convolutional network (GCN) models \cite{c:112,c:113}.
Therefore, it can be seen that these graph-based model only considers the mention words information in the process of constructing the document-graph and inference.

Unfortunately, clue words outside the mentions that are ignored by these methods provide important guiding information for the prediction of many complex inter-sentence relations.
For example, as shown in Figure~\ref{fig1}, although \textbf{“episode”} does not belong to any mentions, it provides important clues for identifying the inter-sentential relations: (\textit{Blackadders, country of origin, United Kingdom}), (\textit{Blackadders, creator, Richard Curtis}) and (\textit{Blackadders, characters, Blackadder}).
Therefore, it is crucial to make full use of clue words outside the mentions in the reasoning process for document-level RE.

In this paper, we treat graph-based document-level RE models as a encoder-decoder framework, in which the encoder is usually a pre-trained language model for extracting contextual embeddings of mentions or entities, and the decoder is a GNN model for inference, called Inference Decoder (\textbf{I-Decoder}).
Similarly, in machine translation tasks, machine translation models based on encoder-decoder framework usually capture the clue information of source language through a decoder-to-encoder attention mechanism to improve translation quality.
Inspired by this, we propose a novel graph-based document-level RE model \textbf{NC-DRE} which introduces a decoder-to-encoder attention mechanism to help the I-Decoder capture clue information, especially non-entity clue information.

Furthermore, to further improve the reasoning ability of the model, inspired by \cite{c:1} that uses transformer architecture to model graphs and achieves better results in graph representation learning compared
to mainstream GNN variants, we use a variant of the Transformer's decoder \cite{c:151} instead of the GNN models as the I-Decoder module for NC-DRE.
In the variant of the Transformer's decoder, we replace the standard Multi-head Self-Attention (\textbf{MSA}) mechanism with a novel Structured-Mask Multi-head Self-Attention (\textbf{SM-MSA}) one that uses the self-attention mechanism to model heterogeneous document-graphs and has better reasoning ability than naive GNNs.
Specifically, in SM-MSA, each head corresponds to a homogeneous sub-graph with a type of edges of the heterogeneous document-graph and performs inference on the sub-graph with a specific mask matrix.

\begin{figure*}[t]
\centering
\includegraphics[width=0.65 \textwidth]{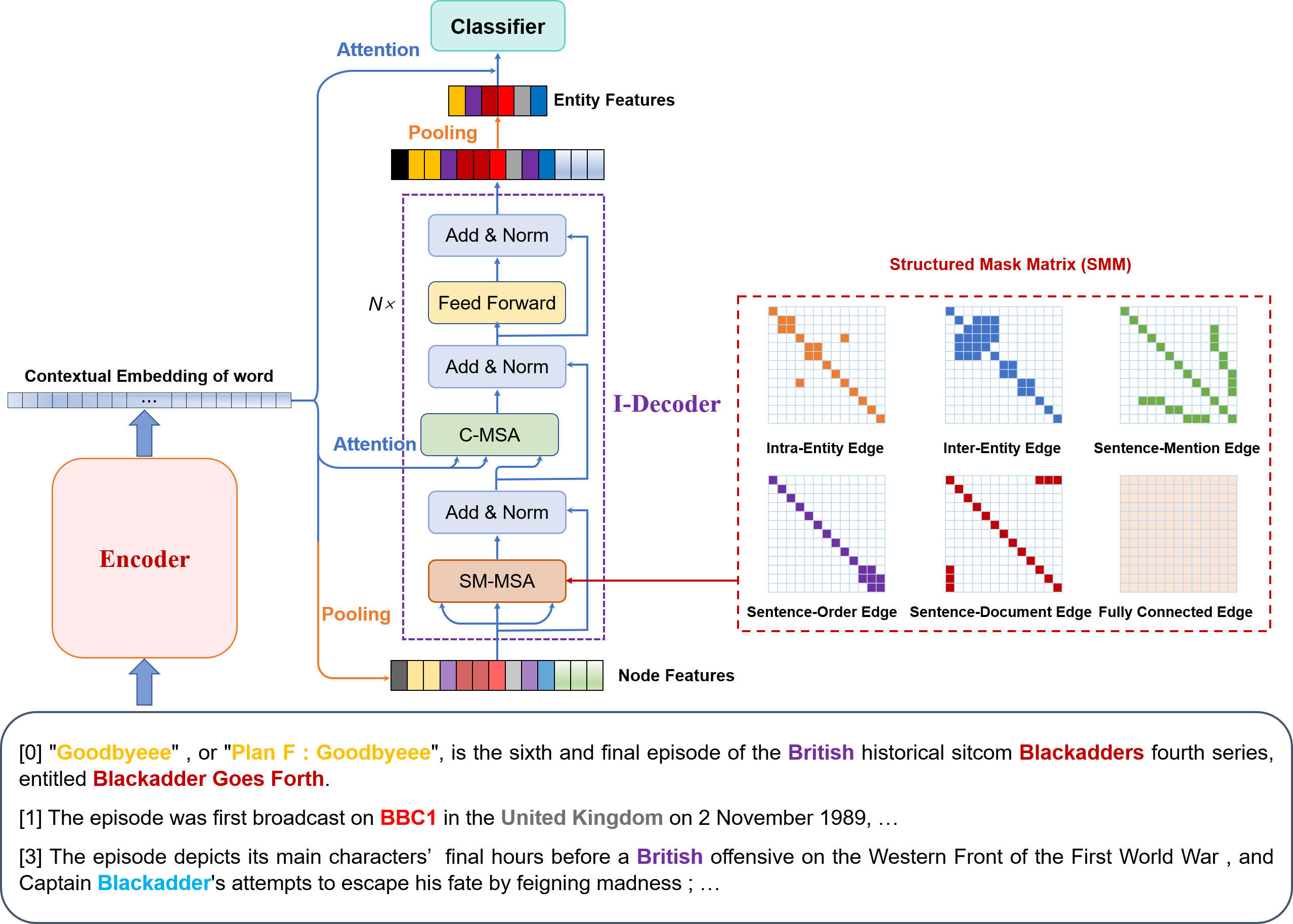} 
\caption{
The architecture of NC-DRE. 
The I-Decoder is a variant of the Transformer's decoder by replacing the standard MSA with our SM-MSA, where C-MSA is a decoder-to-encoder attention mechanism.
SM-MSA contains six heads, and each head corresponds to a specific SMM that is essentially an adjacency matrix of a homogeneous sub-graph in the HMG.
Therefore, NC-DRE can not only complete inference on HMG through SM-MSA but also capture clue information to improve reasoning via C-MSA.
}
\label{fig3}
\end{figure*}

Therefore, as shown in Figure \ref{fig3}, NC-DRE can not only complete inference on heterogeneous document-graphs through SM-MSA but also capture clue information via Cross Multi-head Self-Attention mechanism (\textbf{C-MSA}), i.e. decoder-to-encoder attention mechanism, in Transformer's decoder.
Concretely, NC-DRE first obtains the contextual embedding of each word in the document through the encoder, and then constructs a Heterogeneous Mention-level Graph (\textbf{HMG}), i.e. heterogeneous document-graph, based on the contextual embeddings of words.
Finally, NC-DRE utilizes the SM-MSA to completes the inference on the HMG and captures clue information to improve reasoning  through the C-MSA.

To demonstrate the effectiveness of our NC-DRE model, we conduct comprehensive experiments on three document-level RE datasets, DocRED, CDR, and GDA. 
Our model significantly outperforms the state-of-the-art methods. 
Further analysis demonstrates that our NC-DRE model can effectively capture non-entity clue information during the inference process of document-level RE.
The contributions of our work are summarized as follows:
\begin{itemize}
    \item We propose a novel graph-based document-level RE model NC-DRE, which treats graph-based methods as an encoder-decoder architecture and introduces a C-MSA mechanism to help the decoder capture clue information.
    \item We propose the SM-MSA mechanism to further improve the reasoning ability of our model.
    SM-MSA uses the self-attention mechanism to model MGH and has better reasoning ability than GNN.
    \item We conduct experiments on three public document-level RE datasets, which demonstrate the effectiveness of our NC-DRE model.  
    Meanwhile, our model achieves new state-of-the-art performance on three benchmark datasets.
\end{itemize}


\section{Methodology}
In this section, we introduce in detail our NC-DRE model. 
As shown in Figure~\ref{fig3}, our NC-DRE model consists of three modules, i.e.,  encoder module, I-Decoder module, and classifier module.
We first describe the encoder module and the HMG in Section~\ref{sec2.1}, then introduce our core module, I-Docoder, in Section ~\ref{sec2.2}, finally we describe classifier module and loss function in Section~\ref{sec2.3}.

\subsection{Encoder Module and HMG}
\label{sec2.1}
Given document $D=\{w_i\}_{j=1}^{l}$ containing $l$ words, we first mark the mention in the document by inserting special symbols $\left \langle e_{t} \right \rangle$ and $\left \langle \setminus e_{t} \right \rangle$ at the start and end position of the mentions, where $e_{t}$ is the entity type of the mention. 
It is adapted from the entity marker technique \cite{c:106,c:116,c:152}.
Then we feed the adapted document to the pre-trained language model to obtain the contextual embeddings of each word in the document:
\begin{equation}
H=[h_1,...,h_l] = BERT([w_1,...,w_l]) .
\end{equation}


Then, we construct an HMG based on the contextual embeddings of words.
Concretely, An HMG contains three types of nodes, i.e.,  \textbf{Mention Nodes}, \textbf{Sentence Nodes}, and \textbf{Document Nodes}:

\begin{itemize}
\item We use the contextual embedding of the $\left \langle e_{t} \right \rangle$ located at the start of a mention as the feature vector of the corresponding mention node.
\item For a sentence node, we obtain its feature vector by averaging the contextual embeddings of the words contained in the sentence. 
\item And, we utilize the contextual embedding of $\left \langle CLS \right \rangle$, which is located at the beginning of the document, as the feature vector of the document node.
\end{itemize}

In addition, HMG also contains six types of edges:
\begin{itemize}
    \item \textbf{Intra-Entity Edge:} Different mentions of the entity are fully connected with the intra-entity edges. In this way, we can capture the global contextual information of the entity. 
    \item \textbf{Inter-Entity Edge:} Two mentions of different entities are connected with an inter-entity edge if they co-occur in a single sentence. In this way, the interaction between different entities appearing in the same sentence can be modeled.
    \item \textbf{Sentence-Mention Edge:} If a mention is contained by a sentence, the mention and the sentence will be connected by a Sentence-Mention Edge. In this way, mentions can capture the local contextual information from the sentence in which they are located.
    \item \textbf{Sentence-Order Edge:} Two adjacent sentences in the document are connected with a Sentence-Order Edge. In this way, we can model the sentence-level sequence structure.
    \item \textbf{Sentence-Document Edge:} We connect the document node to each sentence node through the Sentence-Document Edge. In this way, we can obtain an informative document representation.
    \item \textbf{Fully Connected Edge:} All nodes in the HMG are fully connected with the Fully Connected Edge. So the interaction between any two nodes in HMG can be accelerated.
\end{itemize}
As shown in Figure~\ref{fig4}, the HMG contains a node feature matrix $X\in R^{N_n \times d}$ and 6 adjacency matrices $E=\left \langle E_1,E_2,E_3,E_4,E_5,E_6 \right \rangle \in R^{6 \times N_n \times N_n}$  corresponding to 6 kinds of edges, where $N_n$ represents the number of nodes in the HMG.


\subsection{Inference Decoder (I-Decoder)}
\label{sec2.2}
As shown in Figure~\ref{fig3}, our I-Docoder is a variant of Transformer's decoder by replacing the MSA with our SM-MSA.
Therefore, all modules in the I-Decoder except SA-MSA are the same as those in the transformer's encoder.
The I-Decoder uses SM-MSA to perform inference on HMG and C-MSA to capture clue information from the document.

\begin{figure}[t]
\centering
\includegraphics[width=1.0 \columnwidth]{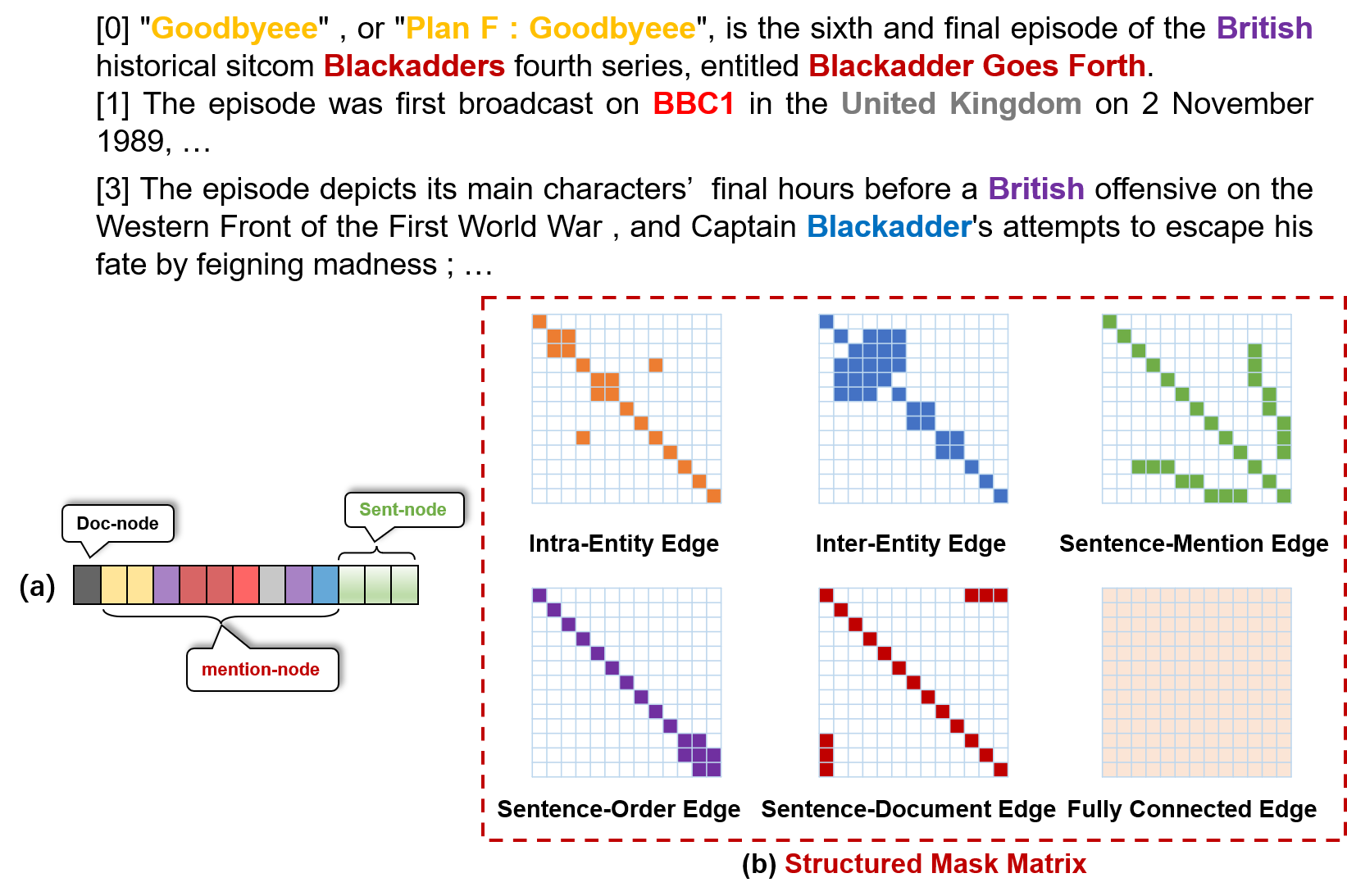} 
\caption{
The illustration of a HMG.
(a) is the node feature matrix which consists of the feature vector of each node in HMG.
(b) shows the adjacency matrices of 6 homogeneous sub-graphs corresponding to the 6 types of edges in the HMG.
In SM-MSA, these adjacency matrices are regarded as mask matrices, called Structured Mask Matrix, to control the information propagation of self-attention on the corresponding sub-graphs.
}
\label{fig4}
\end{figure}

\begin{figure}[t]
\centering
\includegraphics[width=0.85 \columnwidth]{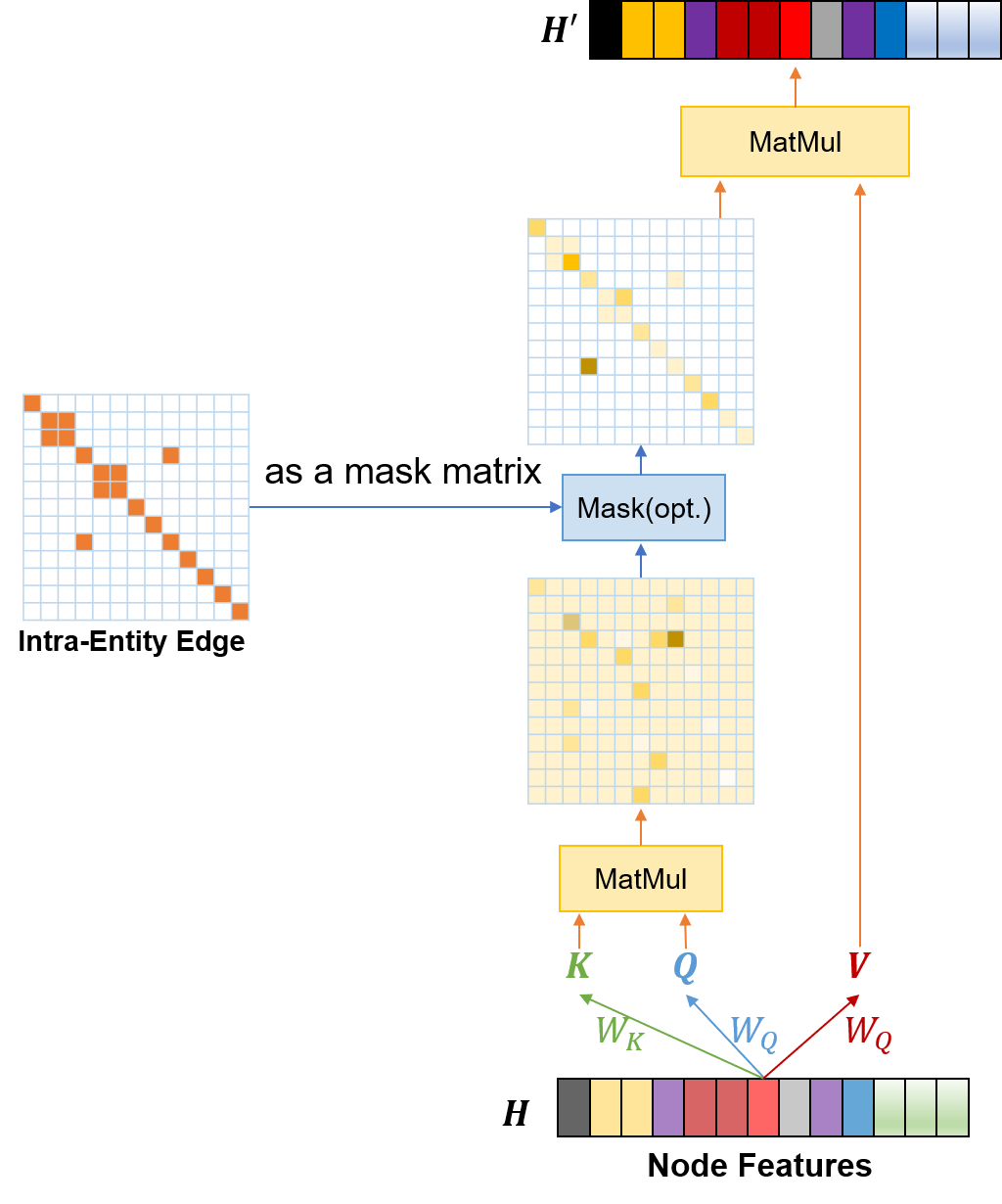}
\caption{
An example illustrating the specific operation of the first head in SM-MSA.
}
\label{fig5}
\end{figure}

To further improve the reasoning ability of our NC-DRE model, inspired by \cite{c:1} that uses the self-attention mechanism to model graph structures and achieves excellent results on graph representation learning tasks, we propose SM-MSA which is a variant of MSA and uses the self-attention mechanism to model HMG.
Specifically, SM-MSA first divides the HMG into 6 homogeneous sub-graphs according to the type of edges and builds a separate adjacency matrix for each sub-graph, as shown in Figure~\ref{fig4}.
In SM-MSA, there are 6 heads and each head corresponds to a homogeneous sub-graph.
Then, to enable each head in SM-MSA to complete inference on the corresponding homogeneous sub-graph, we treat the adjacency matrix of the homogeneous sub-graph as a mask matrix to control the information propagation of self-attention in the head.
Under the restriction of a specific mask matrix, each head in SM-MSA can only transmit information on the corresponding homogeneous sub-graph.
As an example, Figure~\ref{fig5} illustrates the specific operation process of the first head of SM-MSA.

Meanwhile, we capture the clue information for nodes in the HMG from the document via C-MSA.
Since C-MSA can not only capture clue information for the reasoning process but also provide a path for nodes in the HMG to access their own original information, C-MSA can also alleviate the problem of over-smoothing \cite{c:5,c:6,c:7} in graph neural networks (GNN) and enable us to train deeper inference networks.

\subsection{Classification Module}
\label{sec2.3}
Through the I-Decoder module, we obtain a new node feature matrix $X'$ as follows:
\begin{equation}
    X'= I{-}Decoder (X, E, H) ,
\end{equation}

\noindent where $X$ stands for a node feature matrix, $E$ stands for the adjacency matrices and $H$ is the contextual embeddings of each word in the document. 

Then, we use a smooth version of max pooling, i.e. log-sumexp pooling \cite{c:119}, on the feature vectors of the mention nodes to obtain the contextual embedding of the entity $e_i$:
\begin{equation}
h_{e_i}=\log\sum_{j{\in}N_{e_i}}exp(X'_j) ,
\end{equation}
where $N_{e_i}$ is the mention set of the entity $e_i$.

The entity context embedding obtained through I-Decoder inference and pooling operation contains a lot of clue information.
However, for an entity pair, some clue information of the entities may not be relevant, that is, each entity pair may have specific clue information.
Therefore, we generate a specific clue feature $c_{s,o}$ for each entity pair through a filtering operation to remove the irrelevant clue information:
\begin{equation}
	\begin{split}
	    c_{s,o}&=H \cdot a_{s,o}   ,   \\
        a_{s,o}&=softmax(A_s \odot A_o) ,     \\
        A_{*}&=softmax(H \cdot h_{e_*})
	\end{split}
\label{F:4}
\end{equation}
where $*\in \{s,o\}$, $\odot$ refers to element-wise multiplication, and $\cdot$ represents matrix multiplication.

Finally,we use bilinear function as our classifier module as follows:
\begin{equation}
	\begin{split}
	    z_s&=\tanh(W_s[h_{e_s}, c_{s,o}, h_{doc}]) ,\\
	    z_o&=\tanh(W_o[h_{e_o}, c_{s,o}, h_{doc}])  ,\\
	    P(r|e_s,e_o)&=\sigma(z_s^T W_r z_o +b_r) ,
	\end{split}
\end{equation}
where $W_s$, $W_o$, $W_r$, $b_r$ are model parameters, and $h_{doc}$ refers to the feature vector of the document node in $X'$, which can provide document information to assist relation prediction.

To alleviate the problem of unbalanced relationship distribution, we use adaptive-thresholding loss \cite{c:106} as our loss function, which learns an adaptive threshold for each entity pair.
Specifically, a $TH$ class is introduced to separate positive classes and negative classes: positive classes would have higher probabilities than $TH$, and negative classes would have lower probabilities than $TH$.
The adaptive-thresholding loss is formulated as follows:
\begin{equation}
	\begin{split}
	    L_1&={-}\sum_{r{\in}P_D} \log \left(\frac{\exp(logit_r)}{\sum_{r'{\in}\{r,TH\}} exp(logit_{r'})}\right) ,\\
	    L_2&={-}\log \left(\frac{\exp(logit_{TH})}{\sum_{r'{\in}\{N_D,TH\}} exp(logit_{r'})} \right) ,\\
	    L&=L_1+L_2 ,
	\end{split}
\end{equation}
where $P_D$ and $N_D$ are the positive classes set and negative classes set respectively.

\section{Experiments}
\subsection{Datasets}
We conduct experiments on three document-level RE datasets to evaluate our NC-DRE model. 
The statistics of the datasets could be found in Table~\ref{tab6}.
\begin{itemize}
\item \textbf{DocRED} \cite{c:105}:
DocRED is a large-scale human-annotated dataset for document-level RE, which is constructed from Wikipedia and Wikidata. DocRED contains 96 types of relations, 132,275 entities, and 56,354 relationship triples in total. 
In DocRED, more than 40.7\% of relational facts can only be extracted from multiple sentences, and 61.1\% of relational triples require various reasoning skills such as logical inference. 
We follow the standard split of the dataset, 3,053 documents for training, 1,000 for development and, 1,000 for the test.
\item \textbf{CDR} \cite{c:122}:
The Chemical-Disease Reactions dataset (CDR) consists of 1,500 PubMed abstracts, which are split into three equally sized sets for training, development, and testing. 
CDR contains only two types of relationships and is aimed to predict the binary interactions between Chemical and Disease concepts. 
\item \textbf{GDA} \cite{c:123}:
The Gene-Disease Associations dataset (GDA) is a large-scale biomedical dataset, which is constructed from MEDLINE abstracts by the method of distant supervision.
GDA contains 29,192 documents as the training set and 1,000 as the test set. 
GDA is also a binary relation classification task and is aimed to identifies Gene and Disease concepts interactions. 
We follow \cite{c:108} to divide the training set into two parts, 23,353 documents for training and 5,839 for development.
\end{itemize}

\begin{table}[t]
\centering
\setlength{\tabcolsep}{3.5mm}{
\begin{tabular}{lccc}
\toprule
Dataset                     & DocRED     &CDR       &GDA  \\  [2pt] \toprule
Train                     &3053	    &500	    &23353         \\ 
Dev                       &1000	    &500	    &5839       \\
Test                      &1000	    &500	    &1000       \\
Relations                 &97	    &2	    &2        \\
Entities per Doc        &19.5	    &7.6	    &5.4        \\
Mentions per Doc        &26.2	    &19.2	    &18.5        \\ 
Entities per Sent        &3.58	    &2.48	    &2.28        \\     \bottomrule

\end{tabular}}
\caption{\label{tab6} Summary of DocRED, CDR and GDA datasets.
}
\end{table}

\subsection{Experimental Settings}
Our model was implemented based on PyTorch and Huggingface's Transformers \cite{c:120}. 
We use cased BERT-base \cite{c:124} as the encoder on DocRED and cased SciBERT-base \cite{c:126} on CDR and GDA.
We trained our NC-DRE model using AdamW \cite{c:127} with a linear warmup \cite{c:128} for the first 6\% steps followed by a linear decay to 0.
We set the learning rate to $\{2e{-}5, 3e{-}5, 5e{-}5, 1e{-}4, 1e{-}4\}$, and set different learning rates for the encoder module, the I-Decoder module, and the classifier module respectively.
We apply dropout\cite{c:2} between layers with rate 0.1, and clip the gradients of model parameters to a max norm of 1.0. 
By default, we set the number of layers in our I-Decoder module to 4.
All of our hyperparameters were tuned on the development set, some of which are listed in Table~\ref{tab7}.

\subsection{Results on the DocRED Dataset}
On the DocRED Dataset, we choose the following two types of models as the baseline:
\begin{itemize}
    \item \textbf{Graph-based Models}: 
    These models uses the GNN models \cite{c:153,c:117,c:154}  to model long-distance dependencies and complete reasoning on document-level graphs, including GEDA \cite{c:129}, LSR \cite{c:130}, GLRE \cite{c:131}, GAIN \cite{c:118}, and HeterGSAN \cite{c:133}.
    
    \item \textbf{Transformer-based Models}: 
    Considering the transformer architecture can implicitly model long-distance dependencies, this type of method directly use pre-trained language models without graph structures for document-level RE, including $\rm BERT$ \cite{c:133}, BERT-Two-Step \cite{c:134}, HIN-BERT \cite{c:111}, $\rm CorefBERT$ \cite{c:145}, and ATLOP-BERT \cite{c:106}.
\end{itemize}
Moreover, we also compared our model with the SIRE \cite{c:121} model, which is a recently proposed state-of-the-art model. 
SIRE introduces two different methods to represent intra- and inter-sentential relations respectively and design a new and straightforward form of logical reasoning.
Following the previous work \cite{c:105}, we use $F_1$ and Ign$F_1$ as evaluation metrics to evaluate the performance of a model, where Ign$F_1$ denotes the $F_1$ score excluding the relational facts that are shared by the training and dev/test sets. 

\begin{table}[t]
\centering
\setlength{\tabcolsep}{3.0mm}{
\begin{tabular}{lccc}
\toprule
Hyper-parameters                     & DocRED     &CDR       &GDA  \\ 
                            & BERT     &SciBERT       &SciBERT  \\  [2pt] \toprule
Batch size                    &8	    &16	    &16         \\ 
Epoch                       &100	    &20	    &5 \\
lr for encoder                 &3e-5	    &2e-5	    &2e-5        \\
lr for I-Decoder       &1e-4	    &5e-5	    &5e-5         \\
lr for classifier       &2e-4	    &1e-4	    &1e-4         \\
\bottomrule
\end{tabular}}
\caption{\label{tab7} Hyper-parameter Settings.}
\end{table}

\begin{table*}[th]
\centering
\begin{tabular}{lcccccc}
\toprule
Model               & \multicolumn{4}{c}{Dev}                  & \multicolumn{2}{c}{Test} \\
                     & Ign$F_1$         & $F_1$     & Intra-$F1$    & Inter-$F1$   & Ign$F_1$     & $F_1$  \\ [2pt]\toprule
GEDA-BERT \cite{c:129}       & 54.52    & 56.16     & -             & -       & 53.71      & 55.74       \\
LSR-BERT \cite{c:130}        & 52.43    & 59.00        &65.26          &52.05    & 56.97      & 59.05       \\
GLRE-BERT \cite{c:131}       & -        & -         & -             & -       & 55.40       & 57.40        \\
HeterGSAN-BERT \cite{c:133}  & 58.13    & 60.18     & -             & -        & 57.12      & 59.45       \\ 
GAIN-BERT \cite{c:118}       & 59.14    & 61.22     &67.10          &53.90     & 59.00         & 61.24       \\ [2pt] \toprule
BERT \cite{c:134}            & -        & 54.16     &61.61          &47.15      & -          & 53.20        \\
BERT-Two-Step \cite{c:134}   & -        & 54.42     &61.80          &47.28     & -          & 53.92       \\
HIN-BERT\cite{c:111}         & 54.29       & 56.31     & -             & -         & 53.70       & 55.60        \\
CorefBERT\cite{c:145}        & 55.32       & 57.51     & -             & -         & 54.54      & 56.96       \\
ATLOP-BERT\cite{c:106}       & 59.22       & 61.09     & -             & -         & 59.31      & 61.30       \\ 
SIRE-BERT\cite{c:121}        & 59.82     & 61.60     & 68.07         & 54.01     & 60.18     & 62.05 \\  [2pt] \toprule
      
NC-DRE-BERT         & \textbf{60.84}($\pm$0.18) & \textbf{62.75}($\pm$0.16) & \textbf{68.58}($\pm$0.21) & \textbf{55.46}($\pm$0.10) & \textbf{60.59} & \textbf{62.73} \\ \bottomrule

\end{tabular}
\caption{\label{tab1} 
Performance on the development and test set of DocRED. 
We run experiments 5 times with different random seeds and report the
mean and standard deviation on the development set.
Results of all the baseline models come from \cite{c:106,c:121}. 
The results for test set are obtained by submitting to the official Codalab.
}
\end{table*}

The performances of our NC-DRE model and the baseline models on the DocRED dataset are shown in Table~\ref{tab1}. We follow the reports of ATLOP \cite{c:106} and SIRE \cite{c:121} for the scores of these baseline models. The comparisons among all the models shows that our NC-DRE model outperforms the previous state-of-the-art models by \textbf{1.15/1.02} $F_1$/Ign$F_1$ on the dev set and \textbf{0.68/0.41} $F_1$/Ign$F1$ on the test set. 
This demonstrates that our model has excellent overall performance.
Besides, comparing with the graph-based state-of-the-art model, the NC-DRE model outperforms the GAIN model by \textbf{1.53/1.70} $F_1$/Ign$F_1$ on the dev set and \textbf{1.49/1.59} $F_1$/Ign$F_1$ on the test set. 
This demonstrates that non-entity clue information can significantly improve the reasoning ability of the graph-based models in document-level RE.

In addition, we follow GAIN \cite{c:118} and SIRE\cite{c:121} and report Intra-$F_1$ / Inter-$F_1$ scores in Table~\ref{tab1}, which only consider either intra- or inter-sentence relations respectively. 
Intra-sentence relations could be easily recognized because two related entities appear in a single sentence. 
However, identifying inter-sentence relationships is more difficult because related entities appears in different sentences and inference techniques are often required to identify such relationships.
Therefore, Inter-$F_1$ can better reflect the reasoning ability of the model.
We can observe that NC-DRE-BERT improves the Inter-$F_1$ score by \textbf{1.45} compared with the SIRE model.
The improvement on Inter-$F_1$ demonstrates that our NC-DRE model has excellent inference ability.
Moreover, the improvement on Inter-$F_1$ is greater than that on intra-$F_1$, which shows that the performance improvement of NC-DRE is mainly contributed by the improvement of inter-sentence relations.

\subsection{Results on the Biomedical Datasets}
On the two biomedical datasets, CDR and GDA, we compared our model with a large number of baseline models and the recent state-of-the-art models including BRAN \cite{c:136}, EoG \cite{c:108}, LSR \cite{c:130}, DHG \cite{c:138}, GLRE \cite{c:131}, SciBERT \cite{c:126}, and ATLOP \cite{c:106}.  
And the experimental results are listed in Table~\ref{tab2}.

From Table~\ref{tab2}, we observe that our NC-DRE model respectively gets \textbf{72.05} $F_1$ and \textbf{85.80} $F_1$ scores on the CDR and GDA datasets, which outperforms the recent state-of-the-art model with \textbf{1.65} $F_1$ and \textbf{1.9} $F_1$.
This shows that our model has outstanding generalization ability in the field of biomedicine.

In general, graph-based document-level RE models are difficult to perform well on these two datasets.
On the one hand, CDR and GDA contain only two classes of relations, so little reasoning ability is required. On the other hand, samples in CDR and GDA contain few entities, which results in a small document-graph and limit the ability of GNNs.
However, our NC-DRE model still achieves excellent performance on both datasets.
The possible reason is that our NC-DRE model not only has excellent reasoning ability but also can obtain expressive entity representation through I-Decoder.

\subsection{Ablation Study}
We conducted ablation studies on the development set of the DocRED dataset to illustrate the effectiveness of different modules in our model.
We show the results in Table~\ref{tab3}. 

\textbf{w/o C-MHA} removes the C-MHA module from our model, which leads to a drop of \textbf{1.14} $F_1$ score.
This shows that our C-MSA can effectively capture these clue information to significantly improve the inference ability of the model.
Furthermore, \textbf{w/o C-MHA} improves \textbf{0.26} $F_1$ compared with the GAIN model even without the aid of clue information, and GAIN contains a complex entity-level inference module while our model does not.
This demonstrates that our SM-MSA has better inference ability than the GNN models used by previous methods.

\textbf{w/o SM-MHA} replaces SM-MSA with standard MSA in our I-Dcoder, which results in a drop of \textbf{1.83} $F_1$ score.
This demonstrates that our SM-MSA can significantly improve the reasoning ability of MSA on heterogeneous graphs.

\textbf{w/o I-Decoder} removes the I-Decoder module from our NC-DRE model, which leads to a great drop of \textbf{3.60} $F_1$ score.
This shows that our C-MSA and SM-MSA can greatly improve the performance of our base model.

Observing the results of the above ablation experiments, we can find that our C-MSA and SM-MSA can complementarily improve the performance of document-level RE models.

\begin{table}[]
\centering
\begin{tabular}{p{3.7cm}cc} 
\toprule
Model               & CDR          & GDA         \\ [2pt] \toprule
BRAN \citep{c:136}                & 62.1         & -           \\
EoG \citep{c:108}                & 63.6         & 81.5        \\
LSR \citep{c:130}                & 64.8         & 82.2        \\
DHG \citep{c:138}                & 65.9         & 83.1        \\
GLRE \citep{c:131}                & 68.5         & -           \\
SciBERT \citep{c:126}        & 65.1         & 82.5        \\
ATLOP-SciBERT\citep{c:106}   & 69.4         & 83.9        \\  [2pt] \toprule
NC-DRE-SciBERT         & \textbf{72.05$(\pm0.52)$} & \textbf{85.80$(\pm0.20)$}    \\
\bottomrule
\end{tabular}
\caption{\label{tab2}$F_1$ scores on the test sets of CDR and GDA.
The scores of all the baseline models come from \cite{c:106}.
}
\end{table}

\subsection{Analysis \& Discussion}
In this section, we will further discuss and analyze our NC-DRE model from two aspects: (1) the number of layers in the I-Decoder module, and (2) the inference performance.

Table~\ref{tab4} shows the performance of the NC-DRE model with different number of layers in the I-Decoder module.
We can find that increasing the number of layers from 1 to 2 improves the model performance by \textbf{1.02} $F_1$ score.
There are two possible reasons: (1) increasing the number of layers can improve the multi-hop reasoning ability of our NC-DRE model, (2) the NC-DRE model with more layers can capture more clue information for the nodes in the HMG.
However, the performance of the model is slightly improved by \textbf{0.13} $F_1$ when the number of layers is increased from 2 to 4.
Therefore, a two-layer I-Decoder module is sufficient for general cases.
Furthermore, we find that our NC-DRE model with one layer of I-Decoder module achieves similar performance to the SIRE model, which reflects the powerful reasoning ability of our I-Decoder module.

\begin{table}[]
\centering
\setlength{\tabcolsep}{4.5mm}{
\begin{tabular}{lcc}
\toprule
Model                                   & \multicolumn{2}{c}{Dev}   \\ 
                                        & Ign$F_1$      & $F_1$     \\ [2pt] \toprule
NC-DRE-BERT                     & \textbf{60.84}   & \textbf{62.75}         \\ 
w/o C-MHA                  & 59.66       & 61.48       \\
w/o SM-MHA                  & 59.19       & 60.92       \\
w/o I-Decoder         & 57.18       & 59.15      \\  \bottomrule
\end{tabular}}
\caption{\label{tab3} Ablation study of NC-DRE on the development set of DocRED.
}
\end{table}

\begin{table}[]
\centering
\setlength{\tabcolsep}{5.5mm}{
\begin{tabular}{lcc}
\toprule
Layer-number                           & \multicolumn{2}{c}{Dev}   \\ 
                                        & Ign$F_1$      & $F_1$     \\ [2pt] \toprule
1-Layer                           & 59.68       & 61.6       \\
2-Layer                           & 60.79       &62.62       \\
4-Layer                      & \textbf{60.82}       & \textbf{62.75}        \\ \bottomrule
\end{tabular}}
\caption{\label{tab4} Performance of NC-DRE with different number of layers in the I-Decoder module on the development set of DocRED.}
\end{table}

Due to the over-smoothing problem, GNNs are generally set to two layers to obtain the best performance in document-level RE, e.g. GAIN \cite{c:118}.
However, our NC-DRE model that has I-Decoder with 4 layers can still achieve excellent performance, which shows that our model can well alleviate the problem of over-smoothing.
This may benefit from the C-MSA and residual connection in the I-Decoder.

To evaluate the inference ability of the models, we follow \cite{c:118,c:121} and report Infer-$F_1$ scores in table~\ref{tab5}, which only considers relations that engaged in the relational reasoning process. 
We observe that our DRE-MIR model improves \textbf{1.45} Infer-$F_1$ compared with the GAIN model.
Moreover, removing the C-MSA module from our I-Dcoder results in a performance drop of \textbf{1.23} Infer-$F_1$, which demonstrates that clue information can improve the inference ability of the graph-based model.
Meanwhile, the performance of NC-DRE sharply drops by \textbf{5.11} Infer-$F_1$ score when removing the I-Decoder from our NC-DRE model.
This demonstrates that our I-Decoder module can greatly improve the inference ability of our base model.

\begin{table}[]
\centering
\begin{tabular}{p{3.5cm}ccc} 
\toprule
Model                       & Infer-$F_1$      & P         &R         \\ [2pt] \toprule
GAIN-GloVe             &40.82          & 32.76     &54.14           \\
SIRE-GloVe             & 42.72         & 34.83     &55.22        \\ \toprule
BERT-RE             & 39.62         & 34.12     &47.23        \\
RoBERTa-RE              & 41.78         & 37.97     &46.45        \\
GAIN-$\rm BERT$            & 46.89         & 38.71     &59.45           \\[2pt] \toprule
NC-DRE-$\rm BERT$        &  \textbf{48.34}           &  \textbf{40.23}     & \textbf{60.55}  \\
w/o C-MSA          &47.11     &38.93    &59.65  \\
w/o I-Decoder          &43.23     &37.03    &51.93   \\
\bottomrule
\end{tabular}
\caption{Infer-$F_1$ results on the development set of DocRED. P stands for Precision and R stands for Recall.
Results of all the baseline models come from \cite{c:118,c:121}.
}
\label{tab5}
\end{table}

\begin{figure}[t]
{
\setlength{\fboxsep}{0pt}\colorbox{white!0}{\parbox{0.45\textwidth}{
\colorbox{red!3.959018070250749}{\strut "Goodbyeee"} 
\colorbox{red!1.759018070250749}{\strut ,} 
\colorbox{red!1.759018070250749}{\strut or} 
\colorbox{red!3.059018070250749}{\strut "Plan F : Goodbyeee"} 
\colorbox{red!1.759018070250749}{\strut ,} 
\colorbox{red!1.759018070250749}{\strut is} 
\colorbox{red!1.759018070250749}{\strut the} 
\colorbox{red!1.759018070250749}{\strut sixth} 
\colorbox{red!1.759018070250749}{\strut and}
\colorbox{red!7.759018070250749}{\strut final} 
\colorbox{red!40.759018070250749}{\strut episode} 
\colorbox{red!7.759018070250749}{\strut of} 
\colorbox{red!1.759018070250749}{\strut the} 
\colorbox{red!1.759018070250749}{\strut British} 
\colorbox{red!1.759018070250749}{\strut historical}
\colorbox{red!7.759018070250749}{\strut sitcom} 
\colorbox{red!30.759018070250749}{\strut Blackadders} 
\colorbox{red!1.759018070250749}{\strut fourth} 
\colorbox{red!1.759018070250749}{\strut series}
\colorbox{red!1.759018070250749}{\strut ,} 
\colorbox{red!1.759018070250749}{\strut entitled} 
\colorbox{red!32.759018070250749}{\strut Blackadder} 
\colorbox{red!30.7590180702507499}{\strut Goes} 
\colorbox{red!28.759018070250749}{\strut Forth} 
\colorbox{red!8.759018070250749}{\strut .} 
\colorbox{red!1.759018070250749}{\strut The} 
\colorbox{red!45.759018070250749}{\strut episode} 
\colorbox{red!1.759018070250749}{\strut was} 
\colorbox{red!1.759018070250749}{\strut first} 
\colorbox{red!1.759018070250749}{\strut broadcast} 
\colorbox{red!1.759018070250749}{\strut on} 
\colorbox{red!25.759018070250749}{\strut BBC1} 
\colorbox{red!1.759018070250749}{\strut in} 
\colorbox{red!7.759018070250749}{\strut the}
\colorbox{red!33.759018070250749}{\strut United} 
\colorbox{red!35.759018070250749}{\strut Kingdom} 
\colorbox{red!6.759018070250749}{\strut on} 
\colorbox{red!5.759018070250749}{\strut 2} 
\colorbox{red!6.759018070250749}{\strut November} 
\colorbox{red!4.759018070250749}{\strut 1989}
\colorbox{red!7.759018070250749}{\strut ,} 
\colorbox{red!1.759018070250749}{\strut shortly} 
\colorbox{red!3.759018070250749}{\strut before} 
\colorbox{red!12.759018070250749}{\strut Armistice}
\colorbox{red!10.759018070250749}{\strut Day}  
\colorbox{red!1.759018070250749}{\strut .} 
\colorbox{red!1.759018070250749}{\strut (... 7 sentences ...).} 
}}}
\caption{
The heatmap of the attention scores of entity pair (\textit{Blackadders, United Kingdom}) to words in the document, which shows that our NC-DRE mod can pay attention to clue words outside the mentions.
}
\label{fig2}
\end{figure}

\subsection{Case Study}
We choose the example shown in Figure~\ref{fig1} and conduct a case study to illustrate that the entity pairs generated by our I-Decoder module can pay attention to the clue information outside the mentions.
Figure~\ref{fig2} shows the heatmap of the attention scores of entity pair (\textit{Blackadders, United Kingdom}) to words in the document, i.e., $a_{s,o}$ in Formula (\ref{F:4}). 
In fact, the attention score $a_{s,o}$ could be used to identify the clue words that both entity \textit{Blackadders} and entity \textit{United Kingdom} pay attention to together.

Three phenomena can be observed from Figure~\ref{fig2}. Firstly, the concerning entity pair can attend to their own mentions, such as \textit{Blackadders}, \textit{Blackadder Goes Forth}, and \textit{United Kingdom}.
This demonstrates that entities can access their original information through C-MSA to prevent over-smoothing in the process of inference.
Secondly, the concerning entity pair can pay attention to additional related entities, such as: \textit{BBC1} and \textit{Armistice Day}. Actually, these related entities may provide reasoning paths for predicting the relation between the concerning entity pair, such as $ Blackadders\stackrel{r_1}{\rightarrow}BBC1+BBC1\stackrel{r_2}{\rightarrow}United\ Kingdom \stackrel{Infence}{\Longrightarrow} Blackadders\stackrel{r_3}{\rightarrow}United$ $Kingdom$, where $r_1:original\ network$, $r_2:country$, and $r_3:country\ of\ origin$.
Thirdly, the concerning entity pair can well pay attention to important clue words outside the mentions, e.g. \textit{episode}. This phenomenon intuitively shows that our C-MSA mechanism can effectively capture clue information of the non-entity clue words.


\section{Related Work}
\subsection{Sentence-level RE}
Early research on RE focused on sentence-level RE, which predicts the relationship between two entities in a single sentence.
Many approaches \cite{c:139,c:140,c:142,c:143,c:144,c:145,c:146,c:147,c:148,c:149} have been proven to effectively solve this problem.
Since many relational facts in real applications can only be recognized across sentences, sentence-level RE face an inevitable restriction in practice.

\subsection{Document-level RE}
To solve the limitations of sentence-level RE in reality, a lot of recent work gradually shift their attention to document-level RE.

\noindent \textbf{Graph-based Methods}: 
Since GCN can model long-distance dependence and complete logical reasoning, many excellent results have been achieved by recent graph-based models \cite{c:155,c:108,c:129,c:138,c:150,c:131,c:130,c:118,c:133}.
Specifically, they first build a graph structure from the input document and then apply the GNN to the graph to complete logical reasoning. 
The Global Context-enhanced Graph Convolutional Networks (GCGCN) model was proposed by \cite{c:150}, and uses entities as nodes and context of entity pairs as edges between nodes to capture rich global context information of entities in a document.
The LSR model was proposed by \cite{c:130}. The LSR uses a variant of Matrix-Tree Theorem to generate task-specific dependency structures for capturing non-local interactions between entities. In addition, The LSR dynamically build the latent structure through an iterative refinement strategy, which allow the model to incrementally capture the complex interactions for better multi-hop reasoning.
The Graph Aggregation-and-Inference Network (GAIN) model was proposed by \cite{c:118}. The GAIN first constructs a heterogeneous mention-level graph (hMG) to model complex interaction among different mentions across the document, then apply the path reasoning mechanism on an entity-level graph (EG) to infer relations between entities.
The encoder-classifier-reconstructor (HeterGSAN) model was proposed by \cite{c:133}, which uses a reconstructor to model path dependency between the entity pairs with the ground-truth relationship.

\noindent \textbf{Transformer-based Methods}: 
Because the pre-trained language model based on the transformer architecture can implicitly model long-distance dependence and complete logical reasoning, many recent studies \cite{c:111,c:106} directly applied pre-trained language models for document-level RE. The ATLOP \cite{c:106} model directly uses the pre-trained language model along with adaptive thresholding and localized context pooling  techniques for document-level RE.
In addition, \cite{c:121} proposed the SIRE model, which represents intra- and inter-sentential relations in different ways, and design a new and straightforward form of logical reasoning.

Most of these works focus on the words in the mention to fully capture the entity information, while ignoring the important clue information outside the mention in the document. 
So far, none of the models explicitly captures the clue word information to facilitate reasoning on the document-graph.
In this paper, we treat graph-based document-level RE models as a encoder-decoder framework, and introduce decoder-to-encoder attention mechanism to help the decoder capture more clue information, especially those non-entity clue information.
Furthermore, to further improve the reasoning ability of GNN, we propose the RM-MHA mechanism which models heterogeneous document-graphs more effectively by self-attention mechanism.

Furthermore, our work is inspired by recent standard Transformer
architecture based GNN works on graph representation tasks \cite{c:1,c:3,c:4}.
Dwivedi et al. \cite{c:3} revisit a series of works for Transformer-based GNNs, and propose to use Laplacian eigenvector as positional encoding. 
And, their model GT surpasses baseline GNNs on graph representation task.
A concurrent work \cite{c:4} propose a novel full Laplacian spectrum to learn the position of each node in a graph, and empirically shows better results than GT.
\cite{c:1} proposes a Graformer model and  several simple yet effective structural encoding methods to help Graphormer better model graph-structured data. 
Meanwhile, Graformer attain excellent results on a broad range of graph representation learning tasks. 

However, our MS-MSA directly models sparse heterogeneous graphs instead of using fully connected graphs for approximation, which makes our model more straight and efficient.
Specifically, we first decompose the heterogeneous graph into multiple homogeneous sub-graphs, and model each homogeneous sub-graph with a self-attention mechanism and a specific mask matrix.  


\section{Conclusion \& Future Work}
In this paper, we treat the graph-based models in document-level RE as an encoder-decoder framework and propose a novel graph-based model called NC-DRE for document-level RE, which features two core modules: SM-MSA and C-MSA.
Specifically, SM-MSA uses the self-attention mechanism to model heterogeneous document-graphs and has better reasoning ability than the naive GNNs.
Meanwhile, C-MSA is a decoder-to-encoder attention mechanism that can capture more clue information, especially the non-entity clue information, for improving the reasoning of the model.
We conduct experiments on three public document-level RE datasets and the experimental results demonstrate that
our NC-DRE model significantly outperforms the existing models and yields the new state-of-the-art results on all the datasets.
Moreover, further analysis demonstrates that our NC-DRE model can effectively capture non-entity clue information during the inference process, and has excellent interpretability.
In the future, we will try to use our model for other relation extraction tasks, such as dialogue relation extraction and sentence-level relation extraction.

\clearpage
\bibliographystyle{ACM-Reference-Format}
\bibliography{SA-Transformer.bbl}










\end{document}